%% file: main.tex
\documentclass[sigconf]{acmart}
\input{dat620-setup}

\usepackage{lipsum}
\DeclareUnicodeCharacter{2212}{-}
\begin{document}
\title{Vision Transformer Pruning Via Matrix Decomposition}

\author{Tianyi Sun}
\affiliation{Committee on Computational and Applied Mathematics\\ 
Department of Computer Science \\
Department of Statistics\\
University of Chicago}


\begin{abstract}
\input{tex/abstract}

\end{abstract}


\maketitle


\section{Introduction}
\label{sec:introduction}
\input{tex/introduction}

\section{Method}
\label{sec:method}
\input{tex/method}
\section{Experimental Evaluation}
\label{sec:evaluation}
\input{tex/evaluation}
\section{Conclusion}
\label {sec:conclusion}
\input{tex/conclusion.tex}
\section{Future Work}
\label{sec:further}
\input{tex/further}

\bibliographystyle{ACM-Reference-Format}
\bibliography{sample-bibliography}

\end{document}

%% file: dat620-setup.tex
\setcopyright{none}

\acmDOI{}

\acmISBN{}

\acmConference[Computer Vision]{}{Transformer Pruning}{Matrix Computation}
\acmBooktitle{}
\copyrightyear{First version: Dec. 2021}

%% file: tex/abstract.tex
\textbf{This is a further development of Vision Transformer Pruning \cite{zhu2021vision} via matrix decomposition. The purpose of the Vision Transformer Pruning is to prune the dimension of the linear projection of the dataset by learning their associated importance score in order to reduce the storage, run-time memory, and computational demands. In this paper we further reduce dimension and complexity of the linear projection by implementing and comparing several matrix decomposition methods while preserving the generated important features. 
We end up selected the Singular Value Decomposition as the method to achieve our goal by comparing the original accuracy scores in the original Github repository and the accuracy scores of using those matrix decomposition methods, including Singular Value Decomposition, four versions of QR Decomposition, and LU factorization. 
}

%% file: tex/introduction.tex
Transformer \cite{vaswani2017attention} is widely used in Natural Language Processing tasks\cite{sun2021personal}, \cite{sun2023topological}, \cite{sun2005study} and now it has attracted much attention on computer vision applications \cite{chen2021pretrained}, \cite{Chen2020GenerativePF}, \cite{han2021survey}, \cite{liu2021swin}, such as image classification \cite{dosovitskiy2021image}, \cite{han2021transformer}, \cite{touvron2021training}, object detection \cite{carion2020endtoend}, \cite{zhu2021deformable}, and image segmentation \cite{hu2021istr}, \cite{wang2021maxdeeplab}, \cite{wang2021endtoend}. However, most of the transformer used in the set of computer vision tasks are highly storage, run-time memory, and computational resource demanding. Convolutional neural networks (CNNs) are also sharing the same issues. Developed ways of addressing this issues in CNNs' settings are including low-rank decomposition \cite{denton2014exploiting}, \cite{lee2019learning}, \cite{8478366}, \cite{8099498}, \cite{tai2016convolutional}, quantization \cite{cai2017deep}, \cite{gupta2015deep}, \cite{rastegari2016xnornet}, \cite{yang2020searching}, and network pruning \cite{Hassibi1992SecondOD}, \cite{10.5555/109230.109298}, \cite{tang2021scop}, \cite{Tang2020RebornFP}, \cite{wen2016learning}.

Inspired by those methods, several methods has been developed in the setting of vision transformer. Star-Transformer \cite{guo2019startransformer} changes fully-connected structure to the star-shaped topology. There are some methods focus on compressing and accelerating the transformer for the natural language processing tasks. With the emergence of vision transformers such as ViT \cite{dosovitskiy2021image}, PVT \cite{wang2021pyramid}, and TNT \cite{han2021transformer} an efficient transformer is urgently need for computer vision applications. Vision Transformer Pruning (VTP) \cite{zhu2021vision}, inspired by the pruning shceme in network slimming \cite{liu2017learning}, has proposes an approach of pruning the vision transformer according to the learnable important scores. We add the learning importance scores before the components to be prune and sparsify them by training the network with $L_1$ regulation. The dimension with smaller importance scores will be pruned and the compact network can be obtained. The (VTP) method largely compresses and accelerates the original ViT models. Inspired by the low-rank decomposition \cite{denton2014exploiting}, \cite{lee2019learning}, \cite{8478366}, \cite{8099498}, \cite{tai2016convolutional}  methods for compressing and accelerating CNNs, we proposed a matrix decomposition method building up on the VTP to further reduce the storage, run-time memory, and computational resource demanding.

%% file: tex/method.tex
In this section, we first review the key components of Vision Transformer (ViT) \cite{dosovitskiy2021image} and Vision Transformer Pruning according to important features \cite{zhu2021vision}. Then we introduce matrix decomposition methods that is fit particularly for our case.
\begin{figure}[t]
        \includegraphics[width=0.75\columnwidth]{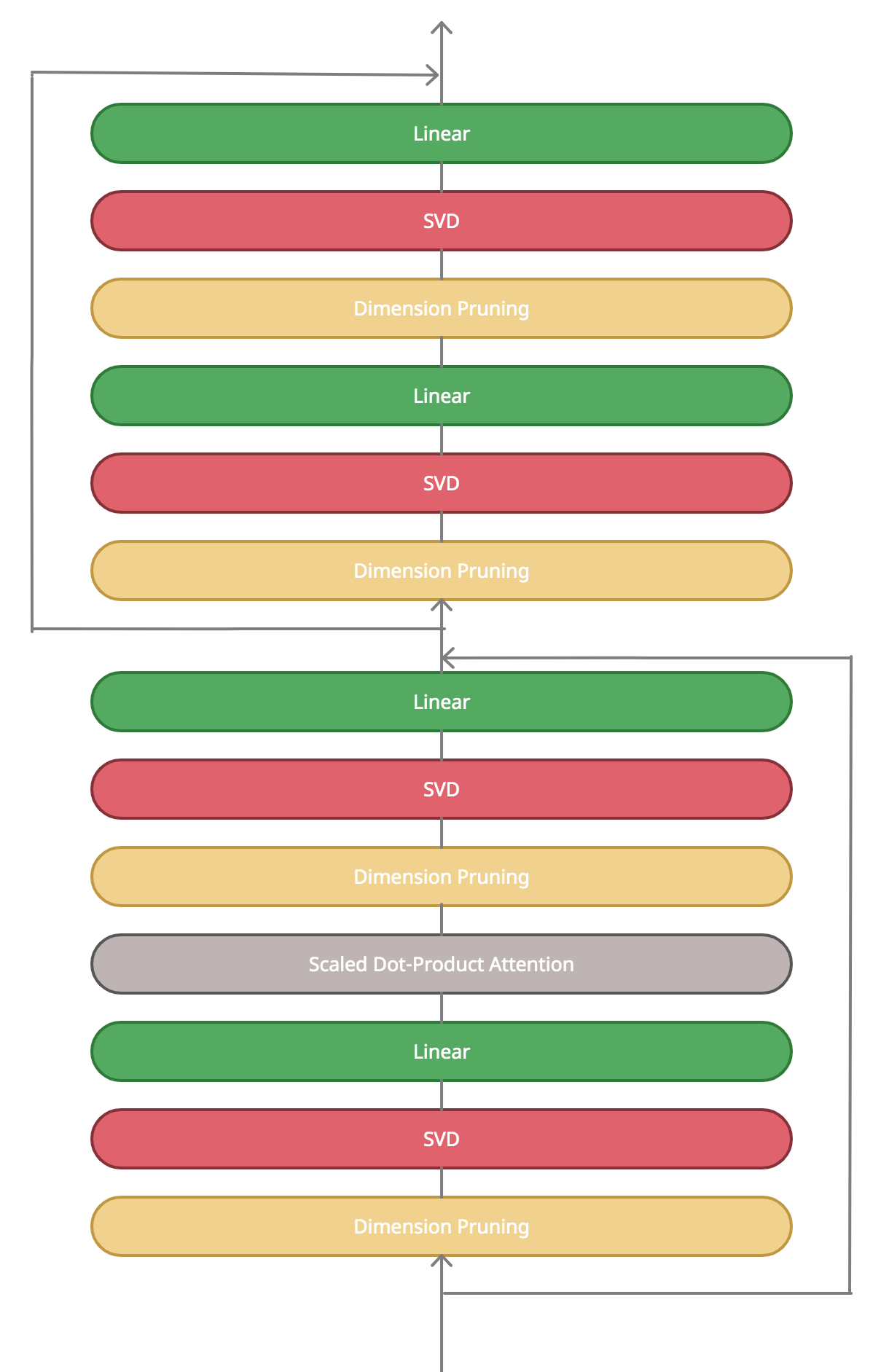}
        \caption{Further prune Vision Transformer via Singular Value  Decomposition (SVD)}
        \label{figVTPMC}
\end{figure}

\subsection{Vision Transformer (ViT)} 
The ViT model includes fully connected layer, Multi-Head Self-Attention(MHSA), linear transformer, Multi-Layer Perception, normalization layer, activation function, and shortcut connection. 
\subsubsection{Fully connected layer}: A fully connected layer is used to transform the input into characters that can be used in the next step, i.e. MHSA. Specifically, it can be denoted as: 
$$I \rightarrow (Q_I,K_I,V_I). $$
The Input $I \in \mathbb{R}^{n \times d}$ is transformed to Query $Q \in \mathbb{R}^{n \times d}$, and Value $V \in \mathbb{R}^{n \times d}$ via fully-connected layers, where $n$ is the number of patches $d$ in the embedded dimension. 

\subsubsection{MHSA}: The self-attention mechanism is utilized to model the relationship between patches. It can be denoted as: $$ Attention(Q_I, K_I, V_I) = Softmax\left(\frac{Q_IK_I^T}{\sqrt{d}}\right)V.$$ 

\subsubsection{Linear Transformer}: A linear transformer is used to transform the output of MHSA. It can be denoted as: 
$$Y = I + f_{out}(Attention(Q_I,K_I,V_I)),$$ where the details of normalization layer and activation function are included in it. 

\subsubsection{Two-layer MLP}: Two feed forward MLP layers can be denoted as: $$ Z = Y + MLP_2(MLP_1(Y)).$$

\subsection{Vision Transformer Pruning} 
The approach of Vision Transformer Pruning is to preserve the generated important features and remove the useless ones from the components of linear projection. 

The pruning process is depending on the score of feature importance. We denote the matrix of feature importance scores is Diag$(a) \in \mathbb{R}^{n \times n} $, where $a \in \mathbb{R}^d$ represents each feature importance score. In order to enforce the sparsity of importance scores, we apply $l_1$ regularization on the importance scores $a$, which is denoted as $\hat{a}$, where $\hat{a} = \lambda||a||_1$ and $\lambda$ is the sparsity hyper-parameter. And optimize it by adding the training objective. We, therefore, obtain the transformer with some important scores near zero. We rank all the values of regularized importance scores in the transformer and obtain a threshold $\tau$ according to the predefined pruning rate. With the threshold, we obtain the discrete $a^* \in \mathbb{R}^{n \times n}$ score, which is denoted as
$$a^* =
\setlength\arraycolsep{1pt}
\left\{
\begin{array}{rcrcrc@{\qquad}l}
1 & \text{ if }\hat{a} \geq \tau\\ 
0 & \text{ if }\hat{a} < \tau.\\ 
\end{array}
\right.
$$
The matrix $X$ after applying the pruning processes, denoted by $$ X^* = X\text{ Diag}(a^*),$$ and removing features with a zero score is transformed to $X^* \in \mathbb{R}^{r \times d}$. This pruning procedure is denoted as $ \text{Prune}(X).$

\subsection{Matrix Decomposition} 

We will introduce an experiential study of Matrix Decomposition methods particularly applied in our case. And a comparison of run-time complexity and the costs of those decomposition. 

\subsubsection{Eigenvalue Value Decomposition (EVD) of $X^*$}

Note that a matrix has an EVD if and only if the matrix is diagonalizable. However, $X^* \in \mathbb{R}^{r \times d}$. When $r \neq d$, the $X^*$ has no EVD. So EVD has limitations while applying in our cases. 

\subsubsection{Singular Value Decomposition (SVD) of $X^*$}

SVD is computed not depending on the dimension of the matrix. SVD can be applied to any matrix. Assume that $X^* \in \mathbb{R}^{m \times n} $ is the matrix after we have done the pruning procedure. Then $X^*$ has a decomposition that $$X^* = U \Sigma V$$ where the left singular vectors $U \in \mathbb{R}^{m \times m} $ and right singular vectors $V \in \mathbb{R}^{n \times n}$ are both unitary matrices, which means $U^*U = I_m = UU^*$ and $V^*V = I_n =VV^*$. $\Sigma \in \mathbb{R}^{m \times n}$ is a diagonal matrix of $X^*$ in the sense that  
$$\sigma_{ij} =
\setlength\arraycolsep{1pt}
\left\{
\begin{array}{rcrcrc@{\qquad}l}
0 & \text{ if } i \neq j \\ 
\mathbb{R}_+ & \text{ if } i= j\\ 
\end{array}
\right.
,$$
where the diagonal values are the set of singular values of the matrix $X^*$. This is the full SVD. We can also remove those singular values that are equal to zero from the matrix, which means $\sigma_{ii}=0$ for $i \in \text{max}\{m,n\}$. Then we get the form of $X^*$ that $$X^* = U_r \Sigma_r V_r,$$ where $ U_r$, $\Sigma_r$, and $V_r$ $\in \mathbb{R}^{n \times n}$. This is the condensed SVD. $r$ is the rank of matrix $X^*$. The columns of $ U_r$ forms an orthonormal basis for the image of $X^*$. The columns of $V_r$ forms an orthonormal basis for the coimage of $X^*$. 

\subsubsection{QR Decomposition of $X^*$} 

The QR decomposition method can solve many problems on the SVD list. The QR decomposition is cheaper to compute in the sense that there are direct algorithms for computing QR and complete orthogonal decomposition in a finite number of arithmetic steps. There are several versions of QR decomposition. 

\begin{itemize}
\item \emph{Full QR Decomposition of $X^*$}: For any matrix of $X^* \in \mathbb{R}^{m\times n} $, where $n\neq m$, there exits an unitary matrix $Q \in \mathbb{R}^{m \times m}$, $Q^*Q = QQ^* = I_n$ and an upper-triangular matrix $R \in \mathbb{R}^{m \times n }$, where $r_{ij}=0$ whenever $i>j$, such that $$X^* = QR = Q\begin{bmatrix}
R_1\\
0
\end{bmatrix}.$$ In which, $R_1 \in \mathbb{R}^{n \times n}$ is an upper-triangular square matrix in general. If $X^*$ has full column rank, which is $rank(X^*) = n$ then $R_1$ is non-singular. 

\item \emph{Reduced QR Decomposition of $X^*$}: For any matrix of $X^* \in \mathbb{R}^{m\times n} $, where $n\neq m$, there exits an orthonormal matrix $Q \in \mathbb{R}^{m \times n}$, $Q^*_1Q_1 = I_n$ but $Q_1Q^*_1 \neq I_m$ unless $m=n$ and an upper-triangular square matrix $R \in \mathbb{R}^{n \times n }$, such that $$X^* = Q_1R_1.$$ In which, $Q_1$ is the first $n$ columns of $Q$ in previous Full QR decomposition. $Q = [ Q_1, Q_2 ]$, where $Q_2 \in \mathbb{R}^{m \times (m-n)}$ is the last $m-n$ columns of $Q$. 
In fact, we can obtain reduced QR decomposition of $X^*$ from the full QR decomposition of $X^*$ by simply multiplying out $$X^* = QR = [Q_1, Q_2]\begin{bmatrix}
R_1\\
0
\end{bmatrix} = Q_1R_1 + Q_20 = Q_1R_1.$$ If $X^*$ has full column rank, which means $rank(A) = n$, then $R_1$ is non-singular. 
\item \emph{Rank-Retaining QR Decomposition of $X^*$}: For matrix of $X^* \in \mathbb{R}^{m\times n} $, where $rank(X^*) = r$, there exists a permutation matrix $	\Pi \in \mathbb{R}^{n \times n}$, a unitary matrix $Q \in \mathbb{R}^{m \times m}$, and a non-singular, upper-triangular square matrix $R_1 \in \mathbb{R}^{r \times r}$, such that $$X^*\Pi = Q \begin{bmatrix}
R_1 & S\\
0 & 0
\end{bmatrix}.$$ In which, $S \in \mathbb{R}^{r \times (n-r)}$ is just some matrix with no special properties. We can also write it as the form that $$X^* = QR\Pi^T=Q\begin{bmatrix}
R_1 & S\\
0 & 0
\end{bmatrix}\Pi^T.$$
\item \emph{Complete Orthogonal Decomposition of $ X^*$}: For matrix of $X^* \in \mathbb{R}^{m\times n}$, where rank$(X^*) = r$, there exists an unitary matrix $Q \in \mathbb{R}^{m \times m}$, an unitary matrix $U \in \mathbb{R}^{n \times n}$, and a non-singular, lower-triangular square matrix $L \in \mathbb{R}^{r \times r}$, such that 
$$X^* = Q\begin{bmatrix}
L & 0\\
0 & 0
\end{bmatrix}U^*.$$ It can be obtained from a full QR decomposition of $\begin{bmatrix}
R^*_1\\
S^*
\end{bmatrix} \in \mathbb{R}^{m \times r}$, which has full column rank, $$\begin{bmatrix}
R^*_1\\
S^*
\end{bmatrix} = Z\begin{bmatrix}
R_2\\
0
\end{bmatrix},$$ where $Z \in \mathbb{R}^{m \times m }$ in unitary and $R_2 \in \mathbb{R}^{r \times r}$ in non-singular, upper-triangular square matrix. Observe the Rank-Retaining QR Decomposition of $X^*$ that $$X^* = Q\begin{bmatrix}
R_1 & S\\
0 & 0
\end{bmatrix}\Pi^T = Q\begin{bmatrix}
R_2^* & 0\\
0 & 0
\end{bmatrix}Z*\Pi^T = Q\begin{bmatrix}
L & 0\\
0 & 0
\end{bmatrix}U^*.$$
\end{itemize}

\subsubsection{LU Decomposition of $X^*$} 
For unstructured matrix $X^*$, the most common method is to obtain a decomposition $X^* = LU$, where $L$ is lower triangular and $U$ is upper triangular. LU decomposition is a sequence of constant LU decomposition. We are not just interested in the final upper triangular matrix, we are also interested in keeping track of all the elimination steps. We will omit this method, because it is intuitively expensive to storage and keep track of the complete L and U factors.

\subsubsection{Cholesky Decomposition of $X^*$} 
The Cholesky Decomposition can be computed directly from the matrix equation $A = R^TR$ where $R$ is upper-triangular. 


The matrix $X^*$ in our vision is dense, the Table \ref{table1} compares the relative merits of normal equations, QR decomposition, and SVD. 
By Table \ref{table1} and an experimental implementation. We finally selected SVD as the method to further prune Vision Transformer. As shown in Figure \ref{figVTPMC}, we apply the SVD operation after each steps of Dimension Pruning operation, which is after pruning operation on all the MHSA and MLP blocks.

\begin{table}[t]
    \def\arraystretch{1.2}
  \begin{tabular}{l|l}
    Merits       &   Evaluation \\ \cline{1-2}
    Accuracy &  Normal Equation $<$ QR Decomposition $<$ SVD   \\
    Speed &  Normal Equation $>$ QR Decomposition $>$ SVD    \\
  \end{tabular}
\caption{A comparison of the relative merits among Normal Equations, QR Decomposition, and SVD.\label{table1}}
\end{table}

%% file: tex/evaluation.tex
In this section, we verify the effectiveness of the purposed methods to further prune the Vision Transformer via Matrix Decomposition on the CIFAR-10 dataset \cite{krizhevsky2010cifar}. 
\subsection{Dataset}

The CIFAR-10 dataset \cite{krizhevsky2010cifar} is divided into five training batches and one test batch, each with $10000$ images. The test batch contains exactly $1000$ randomly-selected images from each class. The training batches contain the remaining images in random order, but some training batches may contain more images from one class than another. Between them, the training batches contain exactly $5000$ images from each class. 

\subsubsection{CIFAR-10.} The CIFAR-10 is labeled subsets of the $80$ million tiny images dataset.  They were collected by Alex Krizhevsky, Vinod Nair, and Geoffrey Hinton. The CIFAR-10 dataset consists of $60000$ images in $10$ classes, with $6000$ images per class. There are $50000$ training images and $10000$ test images.

\subsection{Implementation Details}
We modified the code book\footnote{\url{https://github.com/Cydia2018/ViT-cifar10-pruning}} by adding Matrix Decomposition Methods we mentioned about into the places after doing the Vision Transformer Pruning operation. Since the code book is using the CIFAR$-10$, we do not modify other components of the code book. Figure \ref{table2} shows the result of those implementation. We will discuss it in the next section.

\subsubsection{Baseline} We evaluate developed pruning method via matrix decomposition on a popular vision transformer implementation. In our experiments, a $12-$layer transformer with $10$ heads and $768$ embedding dimensions is evaluated on CIFAR$-10$. For a fair comparison, we utilize the official implementation of CIFAR$-10$. On the CIFAR$-10$, we take the released model of Vision Transformer Pruning as the baseline. We fine-tune the model on the CIFAR$-10$ using batch size $64$ for $30$ epochs. The initial learning rate is set to $6.25 \times 10^{−7}$. Following Deit \cite{touvron2021training}, we use AdamW \cite{loshchilov2018fixing} with cosine learning rate decay strategy to train and fine-tune the models.

\subsubsection{Training with Pruning} Based on the baseline model, we train the vision transformer with $l_1$ regularization using different sparse regularization rates. We select the optimal sparse regularization rate (i.e. $0.0001$) on CIFAR$-10$ and apply it on CIFAR$-10$. The learning rate for training with sparsity is $6.25 \times 10−6$ and the number of epochs is $100$. The other training setting follows the baseline model. After sparsity, we prune the transformer by setting different pruning thresholds and the threshold is computed by the predefined pruning rate. Which is very similar with the original dataset. 
\subsubsection{Fine-tuning} We finetune the pruned transformer with the same optimization setting as in training, except for removing the $l_1$ regularization.
\subsection{Results and Analysis}
In Figure \ref{table2} shows a result of comparison of accuracy after implemented matrix decomposition methods, including SVD, QR decomposition, and LU decomposition. It is easy to see that the SVD performs pretty well comparing with the other two matrix decomposition methods including QR decomposition, and LU decomposition. By experience, the computation is not time consuming. The times of computation is nearly the same except the LU factorization. However, while comparing VTPSVD and VTP, the accuracy score of VTPSVD is not lower than the original VTP. which means that we need to further modify and develop our mathod to achieve a better performance in our case. 
\begin{table}[t]
\footnotesize
\centering
    \begin{tabular*}{\columnwidth}{@{\extracolsep{\fill}}l|llll}
           & VTP & VTPSVD  &  VTPQRD & VTPLUD \\ \cline{1-5}
    ViT patch=2 & 80\%       & 85.0\%        & 71.1\%   & 63.7\%  \\
    ViT patch=4 & 80\%          & 62.8\%        & 68.3\% & 57.1\%  \\
    ViT patch=8 & 30\%         & 53.2\%        & 53.5\%  & 52.3\% \\
    CIFAR-10 & 93\%        & 53.2\%        & 53.5\%   & 49.2\% 
    \end{tabular*}
    \caption{A comparison of accuracy score among Vision Transformer Pruning (VTP), Vision Transformer Pruning with Singular Value Decomposition (VTPSVD), Vision Transformer Pruning with QR Decomposition (VTPQRD), and Vision Transformer Pruning with LU Decomposition (VTPLUD) with different patches. \label{table2}}
\end{table}

%% file: tex/conclusion.tex
In this paper, we build up on and further develop the method of Vision Transformer Pruning by adding a matrix decomposition operation to meet the same goal of reduce the storage, run-time memory, and computational demands. The experiments conducted on CIFAR-10 demonstrate that the pruning with Singular Value Decomposition method can largely reduce the computation costs and storage while maintaining a comparingly high accuracy with some patches. However, this method needs to be further develop to obtain a much robust optimal performance. 

%% file: tex/further.tex
There are several future works that I am planning to do:
\begin{itemize}
\item Test the method on other datasets, i.e. ImageNet \cite{russakovsky2015imagenet}, which is the one used in the original Vision Transformer Pruning \cite{zhu2021vision}.
\item Further develop and fine-tune this method. 
\end{itemize}

%% file: main.bbl

\begin{thebibliography}{37}


\ifx \showCODEN    \undefined \def \showCODEN     #1{\unskip}     \fi
\ifx \showDOI      \undefined \def \showDOI       #1{#1}\fi
\ifx \showISBNx    \undefined \def \showISBNx     #1{\unskip}     \fi
\ifx \showISBNxiii \undefined \def \showISBNxiii  #1{\unskip}     \fi
\ifx \showISSN     \undefined \def \showISSN      #1{\unskip}     \fi
\ifx \showLCCN     \undefined \def \showLCCN      #1{\unskip}     \fi
\ifx \shownote     \undefined \def \shownote      #1{#1}          \fi
\ifx \showarticletitle \undefined \def \showarticletitle #1{#1}   \fi
\ifx \showURL      \undefined \def \showURL       {\relax}        \fi
\providecommand\bibfield[2]{#2}
\providecommand\bibinfo[2]{#2}
\providecommand\natexlab[1]{#1}
\providecommand\showeprint[2][]{arXiv:#2}

\bibitem[\protect\citeauthoryear{Cai, He, Sun, and Vasconcelos}{Cai
  et~al\mbox{.}}{2017}]%
        {cai2017deep}
\bibfield{author}{\bibinfo{person}{Zhaowei Cai}, \bibinfo{person}{Xiaodong He},
  \bibinfo{person}{Jian Sun}, {and} \bibinfo{person}{Nuno Vasconcelos}.}
  \bibinfo{year}{2017}\natexlab{}.
\newblock \bibinfo{title}{Deep Learning with Low Precision by Half-wave
  Gaussian Quantization}.
\newblock
\newblock
\showeprint[arxiv]{cs.CV/1702.00953}


\bibitem[\protect\citeauthoryear{Carion, Massa, Synnaeve, Usunier, Kirillov,
  and Zagoruyko}{Carion et~al\mbox{.}}{2020}]%
        {carion2020endtoend}
\bibfield{author}{\bibinfo{person}{Nicolas Carion}, \bibinfo{person}{Francisco
  Massa}, \bibinfo{person}{Gabriel Synnaeve}, \bibinfo{person}{Nicolas
  Usunier}, \bibinfo{person}{Alexander Kirillov}, {and} \bibinfo{person}{Sergey
  Zagoruyko}.} \bibinfo{year}{2020}\natexlab{}.
\newblock \bibinfo{title}{End-to-End Object Detection with Transformers}.
\newblock
\newblock
\showeprint[arxiv]{cs.CV/2005.12872}


\bibitem[\protect\citeauthoryear{Chen, Wang, Guo, Xu, Deng, Liu, Ma, Xu, Xu,
  and Gao}{Chen et~al\mbox{.}}{2021}]%
        {chen2021pretrained}
\bibfield{author}{\bibinfo{person}{Hanting Chen}, \bibinfo{person}{Yunhe Wang},
  \bibinfo{person}{Tianyu Guo}, \bibinfo{person}{Chang Xu},
  \bibinfo{person}{Yiping Deng}, \bibinfo{person}{Zhenhua Liu},
  \bibinfo{person}{Siwei Ma}, \bibinfo{person}{Chunjing Xu},
  \bibinfo{person}{Chao Xu}, {and} \bibinfo{person}{Wen Gao}.}
  \bibinfo{year}{2021}\natexlab{}.
\newblock \bibinfo{title}{Pre-Trained Image Processing Transformer}.
\newblock
\newblock
\showeprint[arxiv]{cs.CV/2012.00364}


\bibitem[\protect\citeauthoryear{Chen, Radford, Child, Wu, Jun, Luan, and
  Sutskever}{Chen et~al\mbox{.}}{2020}]%
        {Chen2020GenerativePF}
\bibfield{author}{\bibinfo{person}{Mark Chen}, \bibinfo{person}{Alec Radford},
  \bibinfo{person}{Rewon Child}, \bibinfo{person}{Jeffrey Wu},
  \bibinfo{person}{Heewoo Jun}, \bibinfo{person}{David Luan}, {and}
  \bibinfo{person}{Ilya Sutskever}.} \bibinfo{year}{2020}\natexlab{}.
\newblock \bibinfo{title}{Generative Pretraining From Pixels}.
\newblock , \bibinfo{numpages}{1691--1703}~pages.
\newblock


\bibitem[\protect\citeauthoryear{Cun, Denker, and Solla}{Cun
  et~al\mbox{.}}{1990}]%
        {10.5555/109230.109298}
\bibfield{author}{\bibinfo{person}{Yann~Le Cun}, \bibinfo{person}{John~S.
  Denker}, {and} \bibinfo{person}{Sara~A. Solla}.}
  \bibinfo{year}{1990}\natexlab{}.
\newblock \bibinfo{title}{Optimal Brain Damage}.
\newblock , \bibinfo{numpages}{598–605}~pages.
\newblock


\bibitem[\protect\citeauthoryear{Denton, Zaremba, Bruna, LeCun, and
  Fergus}{Denton et~al\mbox{.}}{2014}]%
        {denton2014exploiting}
\bibfield{author}{\bibinfo{person}{Emily Denton}, \bibinfo{person}{Wojciech
  Zaremba}, \bibinfo{person}{Joan Bruna}, \bibinfo{person}{Yann LeCun}, {and}
  \bibinfo{person}{Rob Fergus}.} \bibinfo{year}{2014}\natexlab{}.
\newblock \bibinfo{title}{Exploiting Linear Structure Within Convolutional
  Networks for Efficient Evaluation}.
\newblock
\newblock
\showeprint[arxiv]{cs.CV/1404.0736}


\bibitem[\protect\citeauthoryear{Dosovitskiy, Beyer, Kolesnikov, Weissenborn,
  Zhai, Unterthiner, Dehghani, Minderer, Heigold, Gelly, Uszkoreit, and
  Houlsby}{Dosovitskiy et~al\mbox{.}}{2021}]%
        {dosovitskiy2021image}
\bibfield{author}{\bibinfo{person}{Alexey Dosovitskiy}, \bibinfo{person}{Lucas
  Beyer}, \bibinfo{person}{Alexander Kolesnikov}, \bibinfo{person}{Dirk
  Weissenborn}, \bibinfo{person}{Xiaohua Zhai}, \bibinfo{person}{Thomas
  Unterthiner}, \bibinfo{person}{Mostafa Dehghani}, \bibinfo{person}{Matthias
  Minderer}, \bibinfo{person}{Georg Heigold}, \bibinfo{person}{Sylvain Gelly},
  \bibinfo{person}{Jakob Uszkoreit}, {and} \bibinfo{person}{Neil Houlsby}.}
  \bibinfo{year}{2021}\natexlab{}.
\newblock \bibinfo{title}{An Image is Worth 16x16 Words: Transformers for Image
  Recognition at Scale}.
\newblock
\newblock


\bibitem[\protect\citeauthoryear{Guo, Qiu, Liu, Shao, Xue, and Zhang}{Guo
  et~al\mbox{.}}{2019}]%
        {guo2019startransformer}
\bibfield{author}{\bibinfo{person}{Qipeng Guo}, \bibinfo{person}{Xipeng Qiu},
  \bibinfo{person}{Pengfei Liu}, \bibinfo{person}{Yunfan Shao},
  \bibinfo{person}{Xiangyang Xue}, {and} \bibinfo{person}{Zheng Zhang}.}
  \bibinfo{year}{2019}\natexlab{}.
\newblock \bibinfo{title}{Star-Transformer}.
\newblock
\newblock
\showeprint[arxiv]{cs.CL/1902.09113}


\bibitem[\protect\citeauthoryear{Gupta, Agrawal, Gopalakrishnan, and
  Narayanan}{Gupta et~al\mbox{.}}{2015}]%
        {gupta2015deep}
\bibfield{author}{\bibinfo{person}{Suyog Gupta}, \bibinfo{person}{Ankur
  Agrawal}, \bibinfo{person}{Kailash Gopalakrishnan}, {and}
  \bibinfo{person}{Pritish Narayanan}.} \bibinfo{year}{2015}\natexlab{}.
\newblock \bibinfo{title}{Deep Learning with Limited Numerical Precision}.
\newblock
\newblock
\showeprint[arxiv]{cs.LG/1502.02551}


\bibitem[\protect\citeauthoryear{Han, Wang, Chen, Chen, Guo, Liu, Tang, Xiao,
  Xu, Xu, Yang, Zhang, and Tao}{Han et~al\mbox{.}}{2021a}]%
        {han2021survey}
\bibfield{author}{\bibinfo{person}{Kai Han}, \bibinfo{person}{Yunhe Wang},
  \bibinfo{person}{Hanting Chen}, \bibinfo{person}{Xinghao Chen},
  \bibinfo{person}{Jianyuan Guo}, \bibinfo{person}{Zhenhua Liu},
  \bibinfo{person}{Yehui Tang}, \bibinfo{person}{An Xiao},
  \bibinfo{person}{Chunjing Xu}, \bibinfo{person}{Yixing Xu},
  \bibinfo{person}{Zhaohui Yang}, \bibinfo{person}{Yiman Zhang}, {and}
  \bibinfo{person}{Dacheng Tao}.} \bibinfo{year}{2021}\natexlab{a}.
\newblock \bibinfo{title}{A Survey on Vision Transformer}.
\newblock
\newblock
\showeprint[arxiv]{cs.CV/2012.12556}


\bibitem[\protect\citeauthoryear{Han, Xiao, Wu, Guo, Xu, and Wang}{Han
  et~al\mbox{.}}{2021b}]%
        {han2021transformer}
\bibfield{author}{\bibinfo{person}{Kai Han}, \bibinfo{person}{An Xiao},
  \bibinfo{person}{Enhua Wu}, \bibinfo{person}{Jianyuan Guo},
  \bibinfo{person}{Chunjing Xu}, {and} \bibinfo{person}{Yunhe Wang}.}
  \bibinfo{year}{2021}\natexlab{b}.
\newblock \bibinfo{title}{Transformer in Transformer}.
\newblock
\newblock
\showeprint[arxiv]{cs.CV/2103.00112}


\bibitem[\protect\citeauthoryear{Hassibi and Stork}{Hassibi and Stork}{1992}]%
        {Hassibi1992SecondOD}
\bibfield{author}{\bibinfo{person}{Babak Hassibi} {and} \bibinfo{person}{David
  Stork}.} \bibinfo{year}{1992}\natexlab{}.
\newblock \bibinfo{title}{Second order derivatives for network pruning: Optimal
  Brain Surgeon}.
\newblock
\newblock


\bibitem[\protect\citeauthoryear{Hu, Cao, Lu, Zhang, Wang, Li, Huang, Shao, and
  Ji}{Hu et~al\mbox{.}}{2021}]%
        {hu2021istr}
\bibfield{author}{\bibinfo{person}{Jie Hu}, \bibinfo{person}{Liujuan Cao},
  \bibinfo{person}{Yao Lu}, \bibinfo{person}{ShengChuan Zhang},
  \bibinfo{person}{Yan Wang}, \bibinfo{person}{Ke Li}, \bibinfo{person}{Feiyue
  Huang}, \bibinfo{person}{Ling Shao}, {and} \bibinfo{person}{Rongrong Ji}.}
  \bibinfo{year}{2021}\natexlab{}.
\newblock \bibinfo{title}{ISTR: End-to-End Instance Segmentation with
  Transformers}.
\newblock
\newblock
\showeprint[arxiv]{cs.CV/2105.00637}


\bibitem[\protect\citeauthoryear{Krizhevsky, Nair, and Hinton}{Krizhevsky
  et~al\mbox{.}}{2010}]%
        {krizhevsky2010cifar}
\bibfield{author}{\bibinfo{person}{Alex Krizhevsky}, \bibinfo{person}{Vinod
  Nair}, {and} \bibinfo{person}{Geoffrey Hinton}.}
  \bibinfo{year}{2010}\natexlab{}.
\newblock \showarticletitle{Cifar-10}.
\newblock \bibinfo{journal}{\emph{URL http://www. cs. toronto. edu/kriz/cifar.
  html}}  \bibinfo{volume}{5} (\bibinfo{year}{2010}), \bibinfo{pages}{4}.
\newblock


\bibitem[\protect\citeauthoryear{Lee, Kwon, Kim, and Wei}{Lee
  et~al\mbox{.}}{2019}]%
        {lee2019learning}
\bibfield{author}{\bibinfo{person}{Dongsoo Lee}, \bibinfo{person}{Se~Jung
  Kwon}, \bibinfo{person}{Byeongwook Kim}, {and} \bibinfo{person}{Gu-Yeon
  Wei}.} \bibinfo{year}{2019}\natexlab{}.
\newblock \bibinfo{title}{Learning Low-Rank Approximation for CNNs}.
\newblock
\newblock
\showeprint[arxiv]{cs.LG/1905.10145}


\bibitem[\protect\citeauthoryear{Lin, Ji, Chen, Tao, and Luo}{Lin
  et~al\mbox{.}}{2019}]%
        {8478366}
\bibfield{author}{\bibinfo{person}{Shaohui Lin}, \bibinfo{person}{Rongrong Ji},
  \bibinfo{person}{Chao Chen}, \bibinfo{person}{Dacheng Tao}, {and}
  \bibinfo{person}{Jiebo Luo}.} \bibinfo{year}{2019}\natexlab{}.
\newblock \showarticletitle{Holistic CNN Compression via Low-Rank Decomposition
  with Knowledge Transfer}.
\newblock \bibinfo{journal}{\emph{IEEE Transactions on Pattern Analysis and
  Machine Intelligence}} \bibinfo{volume}{41}, \bibinfo{number}{12}
  (\bibinfo{year}{2019}), \bibinfo{pages}{2889--2905}.
\newblock


\bibitem[\protect\citeauthoryear{Liu, Li, Shen, Huang, Yan, and Zhang}{Liu
  et~al\mbox{.}}{2017}]%
        {liu2017learning}
\bibfield{author}{\bibinfo{person}{Zhuang Liu}, \bibinfo{person}{Jianguo Li},
  \bibinfo{person}{Zhiqiang Shen}, \bibinfo{person}{Gao Huang},
  \bibinfo{person}{Shoumeng Yan}, {and} \bibinfo{person}{Changshui Zhang}.}
  \bibinfo{year}{2017}\natexlab{}.
\newblock \bibinfo{title}{Learning Efficient Convolutional Networks through
  Network Slimming}.
\newblock
\newblock
\showeprint[arxiv]{cs.CV/1708.06519}


\bibitem[\protect\citeauthoryear{Liu, Lin, Cao, Hu, Wei, Zhang, Lin, and
  Guo}{Liu et~al\mbox{.}}{2021}]%
        {liu2021swin}
\bibfield{author}{\bibinfo{person}{Ze Liu}, \bibinfo{person}{Yutong Lin},
  \bibinfo{person}{Yue Cao}, \bibinfo{person}{Han Hu}, \bibinfo{person}{Yixuan
  Wei}, \bibinfo{person}{Zheng Zhang}, \bibinfo{person}{Stephen Lin}, {and}
  \bibinfo{person}{Baining Guo}.} \bibinfo{year}{2021}\natexlab{}.
\newblock \bibinfo{title}{Swin Transformer: Hierarchical Vision Transformer
  using Shifted Windows}.
\newblock
\newblock
\showeprint[arxiv]{cs.CV/2103.14030}


\bibitem[\protect\citeauthoryear{Loshchilov and Hutter}{Loshchilov and
  Hutter}{2018}]%
        {loshchilov2018fixing}
\bibfield{author}{\bibinfo{person}{Ilya Loshchilov} {and}
  \bibinfo{person}{Frank Hutter}.} \bibinfo{year}{2018}\natexlab{}.
\newblock \bibinfo{title}{Fixing Weight Decay Regularization in Adam}.
\newblock
\newblock
\urldef\tempurl%
\url{https://openreview.net/forum?id=rk6qdGgCZ}
\showURL{%
\tempurl}


\bibitem[\protect\citeauthoryear{Rastegari, Ordonez, Redmon, and
  Farhadi}{Rastegari et~al\mbox{.}}{2016}]%
        {rastegari2016xnornet}
\bibfield{author}{\bibinfo{person}{Mohammad Rastegari},
  \bibinfo{person}{Vicente Ordonez}, \bibinfo{person}{Joseph Redmon}, {and}
  \bibinfo{person}{Ali Farhadi}.} \bibinfo{year}{2016}\natexlab{}.
\newblock \bibinfo{title}{XNOR-Net: ImageNet Classification Using Binary
  Convolutional Neural Networks}.
\newblock
\newblock
\showeprint[arxiv]{cs.CV/1603.05279}


\bibitem[\protect\citeauthoryear{Russakovsky, Deng, Su, Krause, Satheesh, Ma,
  Huang, Karpathy, Khosla, Bernstein, Berg, and Fei-Fei}{Russakovsky
  et~al\mbox{.}}{2015}]%
        {russakovsky2015imagenet}
\bibfield{author}{\bibinfo{person}{Olga Russakovsky}, \bibinfo{person}{Jia
  Deng}, \bibinfo{person}{Hao Su}, \bibinfo{person}{Jonathan Krause},
  \bibinfo{person}{Sanjeev Satheesh}, \bibinfo{person}{Sean Ma},
  \bibinfo{person}{Zhiheng Huang}, \bibinfo{person}{Andrej Karpathy},
  \bibinfo{person}{Aditya Khosla}, \bibinfo{person}{Michael Bernstein},
  \bibinfo{person}{Alexander~C. Berg}, {and} \bibinfo{person}{Li Fei-Fei}.}
  \bibinfo{year}{2015}\natexlab{}.
\newblock \bibinfo{title}{ImageNet Large Scale Visual Recognition Challenge}.
\newblock
\newblock
\showeprint[arxiv]{cs.CV/1409.0575}


\bibitem[\protect\citeauthoryear{Sun}{Sun}{2021}]%
        {sun2021personal}
\bibfield{author}{\bibinfo{person}{Tianyi Sun}.}
  \bibinfo{year}{2021}\natexlab{}.
\newblock \bibinfo{title}{How Personal Perceptions Of COVID-19 Have Changed
  Over Time}.
\newblock
\newblock


\bibitem[\protect\citeauthoryear{Sun and Gini}{Sun and Gini}{2021}]%
        {sun2005study}
\bibfield{author}{\bibinfo{person}{Tianyi Sun} {and} \bibinfo{person}{Maria
  Gini}.} \bibinfo{year}{2021}\natexlab{}.
\newblock \showarticletitle{Study of Natural Language Understanding}.
\newblock \bibinfo{journal}{\emph{Journal of Chang Chun Teachers College}}
  \bibinfo{volume}{47} (\bibinfo{year}{2021}).
\newblock


\bibitem[\protect\citeauthoryear{Sun and Nelson}{Sun and Nelson}{2023}]%
        {sun2023topological}
\bibfield{author}{\bibinfo{person}{Tianyi Sun} {and} \bibinfo{person}{Bradley
  Nelson}.} \bibinfo{year}{2023}\natexlab{}.
\newblock \bibinfo{title}{Topological Interpretations of GPT-3}.
\newblock
\newblock
\showeprint[arxiv]{cs.CL/2308.03565}


\bibitem[\protect\citeauthoryear{Tai, Xiao, Zhang, Wang, and E}{Tai
  et~al\mbox{.}}{2016}]%
        {tai2016convolutional}
\bibfield{author}{\bibinfo{person}{Cheng Tai}, \bibinfo{person}{Tong Xiao},
  \bibinfo{person}{Yi Zhang}, \bibinfo{person}{Xiaogang Wang}, {and}
  \bibinfo{person}{Weinan E}.} \bibinfo{year}{2016}\natexlab{}.
\newblock \bibinfo{title}{Convolutional neural networks with low-rank
  regularization}.
\newblock
\newblock
\showeprint[arxiv]{cs.LG/1511.06067}


\bibitem[\protect\citeauthoryear{Tang, Wang, Xu, Tao, Xu, Xu, and Xu}{Tang
  et~al\mbox{.}}{2021}]%
        {tang2021scop}
\bibfield{author}{\bibinfo{person}{Yehui Tang}, \bibinfo{person}{Yunhe Wang},
  \bibinfo{person}{Yixing Xu}, \bibinfo{person}{Dacheng Tao},
  \bibinfo{person}{Chunjing Xu}, \bibinfo{person}{Chao Xu}, {and}
  \bibinfo{person}{Chang Xu}.} \bibinfo{year}{2021}\natexlab{}.
\newblock \bibinfo{title}{SCOP: Scientific Control for Reliable Neural Network
  Pruning}.
\newblock
\newblock
\showeprint[arxiv]{cs.CV/2010.10732}


\bibitem[\protect\citeauthoryear{Tang, You, Xu, Han, Qian, Shi, Xu, and
  Zhang}{Tang et~al\mbox{.}}{2020}]%
        {Tang2020RebornFP}
\bibfield{author}{\bibinfo{person}{Yehui Tang}, \bibinfo{person}{Shan You},
  \bibinfo{person}{Chang Xu}, \bibinfo{person}{Jin Han}, \bibinfo{person}{Chen
  Qian}, \bibinfo{person}{Boxin Shi}, \bibinfo{person}{Chao Xu}, {and}
  \bibinfo{person}{Changshui Zhang}.} \bibinfo{year}{2020}\natexlab{}.
\newblock \showarticletitle{Reborn Filters: Pruning Convolutional Neural
  Networks with Limited Data}.
\newblock \bibinfo{journal}{\emph{Proceedings of the AAAI Conference on
  Artificial Intelligence}} \bibinfo{volume}{34}, \bibinfo{number}{04}
  (\bibinfo{year}{2020}), \bibinfo{pages}{5972--5980}.
\newblock


\bibitem[\protect\citeauthoryear{Touvron, Cord, Douze, Massa, Sablayrolles, and
  Jégou}{Touvron et~al\mbox{.}}{2021}]%
        {touvron2021training}
\bibfield{author}{\bibinfo{person}{Hugo Touvron}, \bibinfo{person}{Matthieu
  Cord}, \bibinfo{person}{Matthijs Douze}, \bibinfo{person}{Francisco Massa},
  \bibinfo{person}{Alexandre Sablayrolles}, {and} \bibinfo{person}{Hervé
  Jégou}.} \bibinfo{year}{2021}\natexlab{}.
\newblock \bibinfo{title}{Training data-efficient image transformers \&
  distillation through attention}.
\newblock
\newblock
\showeprint[arxiv]{cs.CV/2012.12877}


\bibitem[\protect\citeauthoryear{Vaswani, Shazeer, Parmar, Uszkoreit, Jones,
  Gomez, Kaiser, and Polosukhin}{Vaswani et~al\mbox{.}}{2017}]%
        {vaswani2017attention}
\bibfield{author}{\bibinfo{person}{Ashish Vaswani}, \bibinfo{person}{Noam
  Shazeer}, \bibinfo{person}{Niki Parmar}, \bibinfo{person}{Jakob Uszkoreit},
  \bibinfo{person}{Llion Jones}, \bibinfo{person}{Aidan~N. Gomez},
  \bibinfo{person}{Lukasz Kaiser}, {and} \bibinfo{person}{Illia Polosukhin}.}
  \bibinfo{year}{2017}\natexlab{}.
\newblock \bibinfo{title}{Attention Is All You Need}.
\newblock
\newblock
\showeprint[arxiv]{cs.CL/1706.03762}


\bibitem[\protect\citeauthoryear{Wang, Zhu, Adam, Yuille, and Chen}{Wang
  et~al\mbox{.}}{2021c}]%
        {wang2021maxdeeplab}
\bibfield{author}{\bibinfo{person}{Huiyu Wang}, \bibinfo{person}{Yukun Zhu},
  \bibinfo{person}{Hartwig Adam}, \bibinfo{person}{Alan Yuille}, {and}
  \bibinfo{person}{Liang-Chieh Chen}.} \bibinfo{year}{2021}\natexlab{c}.
\newblock \bibinfo{title}{MaX-DeepLab: End-to-End Panoptic Segmentation with
  Mask Transformers}.
\newblock
\newblock
\showeprint[arxiv]{cs.CV/2012.00759}


\bibitem[\protect\citeauthoryear{Wang, Xie, Li, Fan, Song, Liang, Lu, Luo, and
  Shao}{Wang et~al\mbox{.}}{2021a}]%
        {wang2021pyramid}
\bibfield{author}{\bibinfo{person}{Wenhai Wang}, \bibinfo{person}{Enze Xie},
  \bibinfo{person}{Xiang Li}, \bibinfo{person}{Deng-Ping Fan},
  \bibinfo{person}{Kaitao Song}, \bibinfo{person}{Ding Liang},
  \bibinfo{person}{Tong Lu}, \bibinfo{person}{Ping Luo}, {and}
  \bibinfo{person}{Ling Shao}.} \bibinfo{year}{2021}\natexlab{a}.
\newblock \bibinfo{title}{Pyramid Vision Transformer: A Versatile Backbone for
  Dense Prediction without Convolutions}.
\newblock
\newblock
\showeprint[arxiv]{cs.CV/2102.12122}


\bibitem[\protect\citeauthoryear{Wang, Xu, Wang, Shen, Cheng, Shen, and
  Xia}{Wang et~al\mbox{.}}{2021b}]%
        {wang2021endtoend}
\bibfield{author}{\bibinfo{person}{Yuqing Wang}, \bibinfo{person}{Zhaoliang
  Xu}, \bibinfo{person}{Xinlong Wang}, \bibinfo{person}{Chunhua Shen},
  \bibinfo{person}{Baoshan Cheng}, \bibinfo{person}{Hao Shen}, {and}
  \bibinfo{person}{Huaxia Xia}.} \bibinfo{year}{2021}\natexlab{b}.
\newblock \bibinfo{title}{End-to-End Video Instance Segmentation with
  Transformers}.
\newblock
\newblock
\showeprint[arxiv]{cs.CV/2011.14503}


\bibitem[\protect\citeauthoryear{Wen, Wu, Wang, Chen, and Li}{Wen
  et~al\mbox{.}}{2016}]%
        {wen2016learning}
\bibfield{author}{\bibinfo{person}{Wei Wen}, \bibinfo{person}{Chunpeng Wu},
  \bibinfo{person}{Yandan Wang}, \bibinfo{person}{Yiran Chen}, {and}
  \bibinfo{person}{Hai Li}.} \bibinfo{year}{2016}\natexlab{}.
\newblock \bibinfo{title}{Learning Structured Sparsity in Deep Neural
  Networks}.
\newblock
\newblock
\showeprint[arxiv]{cs.NE/1608.03665}


\bibitem[\protect\citeauthoryear{Yang, Wang, Han, Xu, Xu, Tao, and Xu}{Yang
  et~al\mbox{.}}{2020}]%
        {yang2020searching}
\bibfield{author}{\bibinfo{person}{Zhaohui Yang}, \bibinfo{person}{Yunhe Wang},
  \bibinfo{person}{Kai Han}, \bibinfo{person}{Chunjing Xu},
  \bibinfo{person}{Chao Xu}, \bibinfo{person}{Dacheng Tao}, {and}
  \bibinfo{person}{Chang Xu}.} \bibinfo{year}{2020}\natexlab{}.
\newblock \bibinfo{title}{Searching for Low-Bit Weights in Quantized Neural
  Networks}.
\newblock
\newblock
\showeprint[arxiv]{cs.CV/2009.08695}


\bibitem[\protect\citeauthoryear{Yu, Liu, Wang, and Tao}{Yu
  et~al\mbox{.}}{2017}]%
        {8099498}
\bibfield{author}{\bibinfo{person}{Xiyu Yu}, \bibinfo{person}{Tongliang Liu},
  \bibinfo{person}{Xinchao Wang}, {and} \bibinfo{person}{Dacheng Tao}.}
  \bibinfo{year}{2017}\natexlab{}.
\newblock \bibinfo{title}{On compressing deep models by low rank and sparse
  decomposition}.
\newblock , \bibinfo{numpages}{7370--7379}~pages.
\newblock


\bibitem[\protect\citeauthoryear{Zhu, Tang, and Han}{Zhu
  et~al\mbox{.}}{2021b}]%
        {zhu2021vision}
\bibfield{author}{\bibinfo{person}{Mingjian Zhu}, \bibinfo{person}{Yehui Tang},
  {and} \bibinfo{person}{Kai Han}.} \bibinfo{year}{2021}\natexlab{b}.
\newblock \bibinfo{title}{Vision transformer pruning}.
\newblock
\newblock


\bibitem[\protect\citeauthoryear{Zhu, Su, Lu, Li, Wang, and Dai}{Zhu
  et~al\mbox{.}}{2021a}]%
        {zhu2021deformable}
\bibfield{author}{\bibinfo{person}{Xizhou Zhu}, \bibinfo{person}{Weijie Su},
  \bibinfo{person}{Lewei Lu}, \bibinfo{person}{Bin Li},
  \bibinfo{person}{Xiaogang Wang}, {and} \bibinfo{person}{Jifeng Dai}.}
  \bibinfo{year}{2021}\natexlab{a}.
\newblock \bibinfo{title}{Deformable DETR: Deformable Transformers for
  End-to-End Object Detection}.
\newblock
\newblock
\showeprint[arxiv]{cs.CV/2010.04159}


\end{thebibliography}
